\PassOptionsToPackage{hyphens}{url}
\documentclass{article}

\usepackage{arxiv}

\usepackage[utf8]{inputenc}
\usepackage[T1]{fontenc}
\usepackage{hyperref}
\usepackage{url}
\usepackage{booktabs}
\usepackage{amsfonts}
\usepackage{microtype}
\usepackage{graphicx}
\usepackage{natbib}
\usepackage{caption}
\usepackage{array}
\usepackage{placeins}
\newcolumntype{L}[1]{>{\raggedright\arraybackslash}p{#1}}

\title{How Much MRI Preprocessing Is Enough?\\A Cost--Utility Study for Brain MRI Foundation Models}

\author{\textbf{Jiangshuan Pang\textsuperscript{1,2,3} \quad Wangyang Tang\textsuperscript{2,3} \quad Jing Yan\textsuperscript{2,3} \quad Zhixuan Cheng\textsuperscript{2,3}} \\
\textbf{Youzhe He\textsuperscript{2,3} \quad Yiting Deng\textsuperscript{1,2,3} \quad Zhenkun Zhuang\textsuperscript{2,3} \quad Shiping Liu\textsuperscript{2,3,*}} \\
\textsuperscript{1}School of Artificial Intelligence, University of the Chinese Academy of Sciences, Beijing 101408, China\\
\textsuperscript{2}Key Laboratory of Spatial Omics of Zhejiang Province, BGI Research, Hangzhou 310030, China\\
\textsuperscript{3}Key Laboratory of Brain Cell Mapping of Zhejiang Province, BGI Research, Hangzhou 310030, China\\
\textsuperscript{*}Corresponding author: \href{mailto:liushiping@genomics.cn}{liushiping@genomics.cn}}

\date{}

\renewcommand{\shorttitle}{How Much MRI Preprocessing Is Enough?}

\hypersetup{
pdftitle={How Much MRI Preprocessing Is Enough? A Cost--Utility Study for Brain MRI Foundation Models},
pdfsubject={Brain MRI foundation models; preprocessing; self-supervised learning},
pdfauthor={JiangShuan Pang, Wangyang Tang, Jing Yan, Zhixuan Cheng, Youzhe He, Yiting Deng, Zhenkun Zhuang, Shiping Liu},
pdfkeywords={brain MRI, preprocessing, foundation models, self-supervised learning},
}

\begin{document}

\maketitle

\begin{abstract}
MRI preprocessing defines the input distribution seen by brain MRI foundation models, yet it is usually treated as routine data cleaning rather than a modeling choice. We ask how much preprocessing is worth its computational cost for self-supervised 3D MRI pretraining. Keeping the corpus, 3D ViT backbone, masking protocol, and downstream evaluations fixed, we compare a graded P0--P7 preprocessing spectrum for masked autoencoding (MAE) and joint-embedding predictive learning (JEPA) on 20,000 heterogeneous brain MRI volumes, then transfer the encoders to IDH prediction, MCI classification, brain age regression, and GLI/PED tumor segmentation. The results do not support a simple ``more is better'' rule. P0/P1 are numerically unstable, making P2 the lowest-cost feasible level; beyond P2, choosing the best feasible preprocessing level improves aggregate utility by only 3.4 percentage points for MAE and 1.8 percentage points for JEPA, with most paired gains statistically unresolved. Stronger preprocessing is beneficial only in selected regimes: IDH improves modestly, AGE and GLI/PED are often near or best at P2, and MCI shows the clearest empirical P7 gain. Cross-level MCI transfer further shows that much of the P7 advantage can be recovered by applying stronger preprocessing downstream, without requiring P7 throughout pretraining. These findings recast MRI preprocessing as a downstream-aware cost--utility decision rather than a default escalation pipeline.
\end{abstract}

\section{Introduction}

Large-scale self-supervised representation learning has become a central recipe for medical image analysis and medical foundation models \citep{moor2023gmai,rajpurkar2022ai}. Brain MRI models increasingly adopt this recipe through large-scale 3D pretraining and domain-aware representation learning \citep{kim2024domain,dong2025mricore}. Yet the input definition used for pretraining often receives less scrutiny than the architecture, objective, or dataset scale. In practice, MRI preprocessing is not merely a cleaning step. Brain MRI volumes differ in orientation, voxel spacing, field of view, scanner-dependent intensity scale, bias-field artifacts, skull and non-brain tissue content, and anatomical coordinate frame. Preprocessing can reduce these sources of heterogeneity, but it also decides which acquisition cues, subject-specific anatomy, and pathology-related variation remain visible to the model. The preprocessing pipeline therefore becomes part of the representation-learning problem itself.

\begin{figure}[!t]
\centering
\includegraphics[width=\columnwidth]{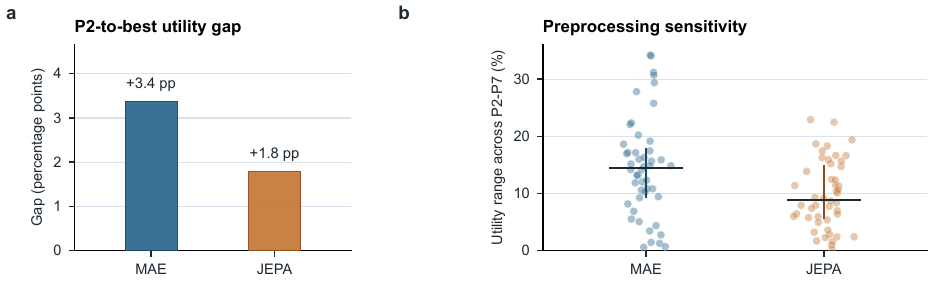}
\caption{Aggregate objective-level evidence for cost--utility and preprocessing robustness. Left: total-utility gap between P2 and the best feasible preprocessing level. Right: preprocessing sensitivity across matched evaluation units.}
\label{fig:objectivegap}
\end{figure}

\begin{figure*}[t]
\centering
\includegraphics[width=\textwidth]{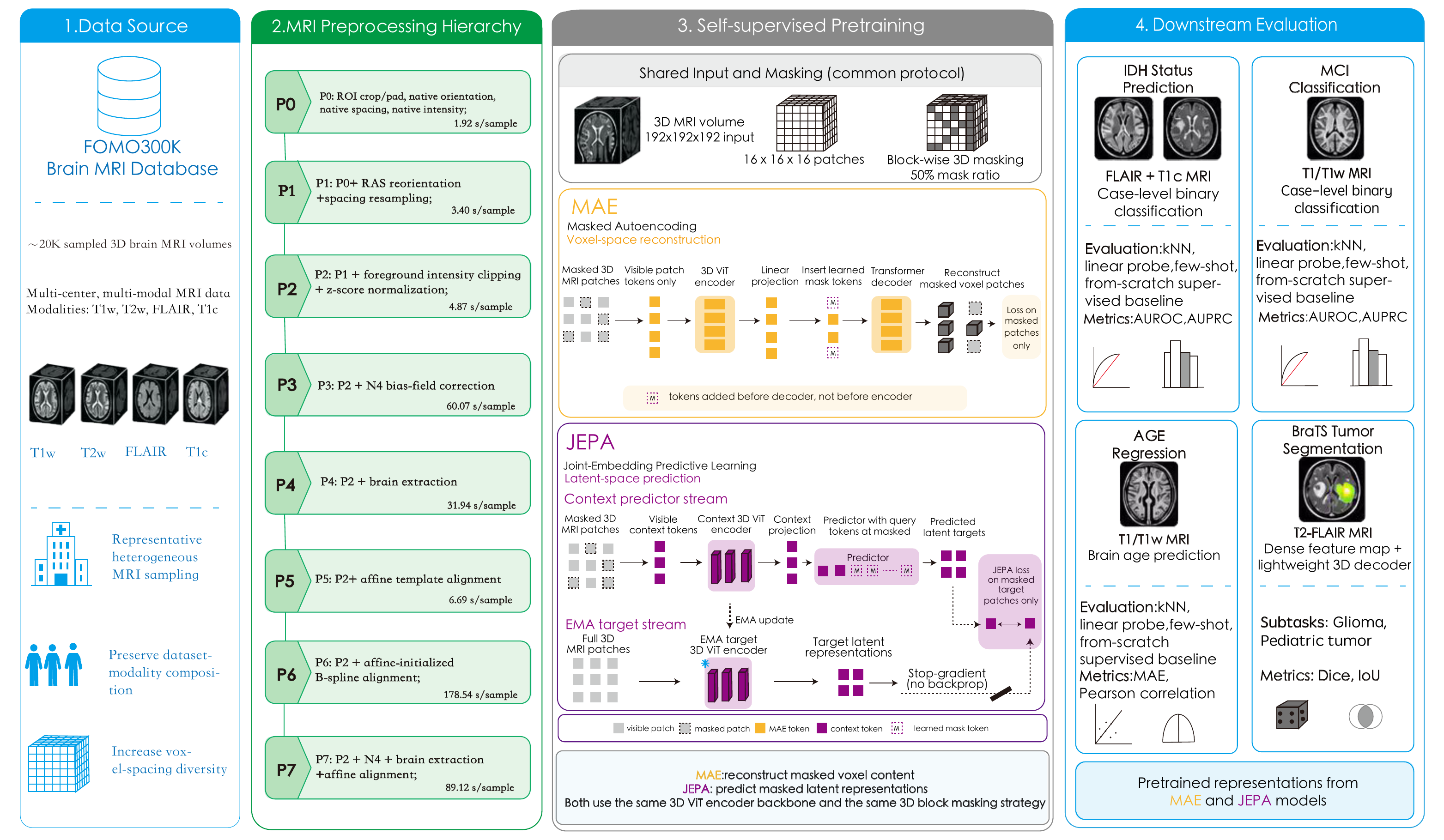}
\caption{Study overview. We sample heterogeneous 3D brain MRI volumes, construct a graded preprocessing hierarchy, pretrain the same 3D ViT encoder with MAE and JEPA under a shared masking protocol, and evaluate the resulting representations on IDH prediction, MCI classification, brain age regression, and GLI/PED BraTS tumor segmentation. The design isolates preprocessing strength as the primary experimental variable while keeping the backbone and downstream protocols fixed.}
\label{fig:overview}
\end{figure*}

This paper asks a direct question: \emph{how much MRI preprocessing is enough for brain MRI foundation models?} We do not propose a new encoder architecture. Instead, we isolate preprocessing strength as the primary experimental variable. We pretrain the same 3D ViT encoder \citep{dosovitskiy2021vit} on the same 20,000-volume brain MRI corpus sampled from FOMO300K \citep{cerri2026large,fomomri2026fomo300k} under a graded sequence of preprocessing definitions. We evaluate two self-supervised objectives, masked autoencoding (MAE) \citep{he2022mae} and joint-embedding predictive learning (JEPA) \citep{assran2023ijepa}, and transfer the resulting encoders to downstream tasks spanning molecular classification, cognitive impairment classification, brain age regression, and dense tumor segmentation. Figure~\ref{fig:overview} summarizes the study design.

The key methodological point is that preprocessing must be evaluated together with cost. In our current pretraining setup, the cheaper P0/P1 pipelines were not usable: their raw or weakly normalized intensity distributions produced NaN gradients and loss collapse rather than stable convergence, with failures occurring before the first epoch completed in both objectives. We interpret this failure primarily as a consequence of insufficient intensity standardization in P0/P1 under the tested 3D ViT and MAE/JEPA training recipe, rather than as evidence that weak spatial preprocessing is universally untrainable. We therefore treat P2, which adds spatial standardization and foreground-based intensity normalization, as the first feasible baseline. This baseline changes the question. The issue is not whether P2 is on the cost--utility frontier---no cheaper feasible pipeline remains---but whether more expensive steps such as N4 correction, skull stripping, affine registration, deformable registration, and combined aggressive standardization produce enough downstream utility to justify leaving P2.

Figure~\ref{fig:objectivegap} gives an aggregate view of this trade-off. We normalize each metric against the best preprocessing level within matched evaluation units, then average metrics within protocols, few-shot shots within the few-shot protocol, protocols within each task, and the four tasks with equal weight. Under this total utility score, replacing P2 with the best feasible preprocessing level improves MAE by only 3.4 percentage points (pp) and JEPA by only 1.8 pp. The matched-unit sensitivity distribution is also lower for JEPA, suggesting greater robustness to preprocessing choices, although individual evaluation units can still be sensitive.

Our main findings are:
\begin{itemize}
\item P2 is the lowest-cost feasible preprocessing level. P0/P1 are cheaper but fail with numerical instability, whereas every usable alternative adds preprocessing cost.
\item Heavier preprocessing provides limited overall gain. Replacing P2 with the best feasible level improves aggregate utility by only 3.4 pp for MAE and 1.8 pp for JEPA, and most paired P2-vs-best comparisons remain statistically unresolved.
\item The benefit of stronger preprocessing is task-dependent rather than monotonic. IDH shows modest gains, AGE is usually close to P2, and GLI/PED fixed-input segmentation is best at P2; MCI shows the clearest empirical P7 gain.
\item The MCI gain is largely recoverable at downstream preprocessing time. P2-pretrained checkpoints evaluated on P7-processed MCI data recover 68.6\% (JEPA) and 81.1\% (MAE) of the gap to the matched P7-checkpoint/P7-data reference, whereas P7 checkpoints degrade on P2-processed MCI data.
\end{itemize}

\section{Related Work}

\textbf{Self-supervised vision pretraining.}
Vision self-supervision spans contrastive learning and self-distillation \citep{chen2020simclr,he2020moco,caron2021dino}, masked image modeling \citep{bao2022beit,he2022mae,xie2022simmim}, and latent prediction objectives \citep{assran2023ijepa}. Masked autoencoding learns representations by reconstructing missing image content from visible context, while joint-embedding predictive learning predicts representations of masked regions in latent space, reducing dependence on pixel-level reconstruction targets. These paradigms have been adapted to 3D medical imaging through generic volumetric pretraining \citep{zhou2021modelsgenesis} and transformer-based medical encoders \citep{tang2022swinunetr}, where volumetric context and anatomical structure are central to transfer.

\textbf{Brain MRI foundation models.}
Recent brain MRI representation learning has explored large-scale 3D pretraining and domain-aware objectives for transfer to clinical and demographic tasks \citep{kim2024domain}. Brain- and MRI-specific foundation models now cover segmentation-oriented neuroimage pretraining \citep{cox2024brainsegfounder}, anatomical brain MRI encoders \citep{barbano2026anatcl}, MRI-wide pretraining \citep{dong2025mricore}, and generative multimodal brain imaging \citep{yang2025genbrain}. BrainIAC is a particularly relevant recent example: it uses contrastive self-supervised pretraining with a 3D vision encoder to learn general-purpose representations from multiparametric brain MRI, and evaluates transfer across sequence classification, brain age prediction, MCI classification, IDH prediction, survival prediction, time-to-stroke prediction, and tumor segmentation \citep{tak2026brainiac}. Broader radiology or medical 3D pretrained models, including RadFM \citep{wu2023radfm} and MedicalNet/Med3D \citep{chen2019med3d}, are also common transfer baselines. These studies typically focus on data scale, model capacity, or objective design. Our work is complementary: we ask how the MRI input definition itself shapes the learned representation.

\textbf{MRI preprocessing.}
MRI preprocessing pipelines commonly include reorientation, resampling, intensity normalization \citep{nyul1999standardizing}, N4 bias-field correction \citep{tustison2010n4}, skull stripping \citep{smith2002bet}, and registration to a template. Toolkits such as ANTs \citep{avants2011ants} and FSL \citep{jenkinson2012fsl} make these steps routine. Classical neuroimaging workflows often combine segmentation, bias correction, and spatial normalization for group-level analysis \citep{ashburner2005unified}. However, the benefit of stronger standardization for self-supervised foundation-model transfer is not obvious. Registration may improve anatomical correspondence, but it may also reduce individual variation; skull stripping may remove distracting non-brain tissue, but can fail near pathology; bias correction may stabilize intensities, but is computationally expensive. We therefore study preprocessing strength as an empirical design variable rather than assuming a universally optimal pipeline.

\section{Preprocessing Spectrum}

We define a hierarchy of preprocessing levels, P0--P7, with increasing standardization strength. All feasible levels ultimately produce fixed-size 3D inputs for the same encoder. Panel 2 of Figure~\ref{fig:overview} summarizes the levels and measured mean processing time per volume; the complete image definitions, implementation details, absolute preprocessing times, and P0/P1 failure diagnostics are reported in the supplementary material. P0 and P1 are included as boundary conditions: in our current pretraining implementation, their raw and weakly normalized intensity distributions caused numerical instability, NaN gradients, and loss collapse under the standardized pretraining setup before completing one epoch. They are therefore treated as infeasible, making P2 the lowest-cost feasible level.

This hierarchy is designed to separate distinct sources of standardization. P2 controls orientation, spatial sampling, and foreground intensity scale. P3 isolates bias-field correction. P4 isolates skull stripping. P5 and P6 introduce global and deformable anatomical alignment. P7 combines several aggressive steps. Qualitative examples of the resulting pretraining inputs are provided in the supplementary material. We do not assume that stronger preprocessing is better; rather, we use this spectrum to measure where the cost--utility trade-off changes.

\section{Experimental Setup}

\textbf{Pretraining corpus.}
The pretraining corpus is a 20,000-volume sample from the gated FOMO300K superset \citep{cerri2026large,fomomri2026fomo300k}, with nonzero samples from 26 constituent datasets and spanning common structural sequences including T1w, T2w, FLAIR, and T1c. Sampling preserves dataset and modality composition while maintaining spatial-resolution diversity; the supplementary material reports the source-level composition. The self-supervised pretraining stage does not use downstream labels, fold assignments, validation metrics, or test outcomes; downstream model selection and final reporting are performed only within the fixed evaluation splits described below.

\textbf{Encoder and objectives.}
All models use the same 3D ViT-Base encoder with 12 transformer layers, hidden size 768, 12 attention heads, input size $192^3$, and non-overlapping $16^3$ patches. We evaluate MAE and JEPA under the same preprocessing levels. MAE reconstructs masked voxel patches with a lightweight decoder, while JEPA predicts masked latent representations using an EMA teacher and predictor. Apart from preprocessing, training configurations are fixed within each objective; the masking protocol, objective heads, optimizer settings, precision, seeds, and training length are given in the supplementary material.

\textbf{Downstream evaluation.}
We evaluate four downstream task families:
\begin{itemize}
\item \textbf{IDH classification}: binary glioma IDH status prediction from UCSF-PDGM baseline MRI \citep{calabrese2022ucsfpdgm} using FLAIR and T1c, with 495 cases.
\item \textbf{MCI classification}: OASIS-1-derived cognitive impairment classification from T1/T1w MRI \citep{marcus2007oasis}, with 235 samples.
\item \textbf{AGE regression}: chronological age prediction from T1/T1w MRI pooled from HCP S1200, LONG579, SALD, Calgary, PETfrog, and IXI \citep{hcpS1200,wang2022longitudinal,wei2018sald,reynolds2020calgary,luna2020petfrog,ixiDataset}, with 3578 samples.
\item \textbf{BraTS segmentation}: binary tumor segmentation from T2-FLAIR using GLI and PED subsets \citep{baid2021brats,bratspedsTcia}.
\end{itemize}

For IDH, MCI, and AGE, we use fixed five-fold cross-validation. The held-out fold is used only for testing; a validation set is separated from the remaining samples for hyperparameter selection, early stopping, and classification threshold selection. Encoders are frozen for kNN, linear probe, and few-shot evaluations. IDH and MCI report AUROC and AUPRC as primary classification metrics; AGE reports MAE and Pearson correlation. For BraTS, the encoder is frozen and a lightweight 3D decoder is trained on dense feature maps; Dice and IoU are reported on test cases. The supplementary material provides the full task definitions, split construction, input-consistency rules, and downstream optimization settings.

\section{Results}

The results support a cost-aware view of brain MRI preprocessing. Once a stable minimum input definition is reached, stronger preprocessing is not uniformly rewarded. P2 is the lowest-cost level that supports stable pretraining, and it already preserves most downstream utility for AGE and fixed-input GLI/PED segmentation, with only modest gains left for IDH. MCI shows the clearest empirical exception, but cross-level transfer shows that much of its P7 advantage can be recovered through downstream input standardization rather than by requiring P7 throughout foundation-model pretraining. We present the evidence in four steps: we first verify that the pretrained encoders are competitive, then quantify the P2-centered cost--utility trade-off, analyze the MCI pattern, and examine objective-specific effects.

\subsection{Pretrained Encoders Are Competitive Baselines}

Before interpreting preprocessing cost--utility, we first verify that the representations under analysis are not weak or degenerate checkpoints. Table~\ref{tab:baseline-sanity} compares the best MAE/JEPA model in each representative setting with a randomly initialized ViT trained from scratch and with the strongest available external baseline for that setting. The external baselines include BrainIAC \citep{tak2026brainiac}, GenBrain \citep{yang2025genbrain}, MedicalNet/Med3D \citep{chen2019med3d}, MRI-Core \citep{dong2025mricore}, and RadFM \citep{wu2023radfm}; full task-specific details are provided in the supplementary material.

Across IDH linear probing, MCI linear probing, AGE kNN regression, and GLI/PED macro segmentation, the selected pretrained encoders are competitive rather than merely functional. They outperform ViT-Scratch and achieve better primary metrics than the strongest available external baseline in these representative settings. The clearest margins appear in MCI linear probing, where the pretrained model reaches 0.7533 AUROC compared with 0.6911 for the best external baseline and 0.6656 for scratch, and in IDH linear probing, where it reaches 0.8767 AUROC compared with 0.8547 and 0.7939. For AGE, where lower MAE is better, the pretrained kNN representation obtains 2.4676 MAE, improving over both the strongest external baseline (2.5146) and scratch training (2.6815). The same pattern extends to dense prediction: under the standardized BraTS fixed-input protocol, the GLI/PED macro Dice result shows that the frozen encoder remains a strong segmentation-transfer baseline, while leaving raw-data preprocessing ablation to the controlled P-level experiments.

These comparisons establish the downstream competence of the encoders used in the P-level analysis. They are not the main evidence for the preprocessing claim, which comes from controlled comparisons across P-levels, but they ensure that the following cost--utility results are measured on competitive pretrained representations.

\begin{table*}[!htbp]
\centering
\small
\setlength{\tabcolsep}{6pt}
\begin{tabular}{@{}llrlrlrl@{}}
\toprule
 & & \multicolumn{2}{c}{Ours} & \multicolumn{2}{c}{Best external} & \multicolumn{2}{c}{Scratch} \\
\cmidrule(lr){3-4}\cmidrule(lr){5-6}\cmidrule(l){7-8}
Setting & Metric & Value & Model & Value & Model & Value & Model \\
\midrule
IDH linear & AUROC $\uparrow$ & \textbf{0.8767} & JEPA P6 & 0.8547 & MRI-Core & 0.7939 & ViT P6 \\
MCI linear & AUROC $\uparrow$ & \textbf{0.7533} & JEPA P7 & 0.6911 & MRI-Core & 0.6656 & ViT P2 \\
AGE kNN & MAE $\downarrow$ & \textbf{2.4676} & JEPA P5 & 2.5146 & MRI-Core & 2.6815 & ViT P5 \\
BraTS GLI/PED macro & Dice $\uparrow$ & \textbf{0.8033} & JEPA P2 & 0.7934 & MRI-Core & 0.7411 & ViT \\
\bottomrule
\end{tabular}
\caption{Baseline sanity check against scratch and external baselines. Ours is the best MAE/JEPA checkpoint selected for each setting; Scratch denotes ViT-Scratch. AGE reports MAE (lower is better), whereas the other settings report AUROC or Dice (higher is better).}
\label{tab:baseline-sanity}

\end{table*}

\subsection{P2 Captures Most Utility at the Lowest Feasible Cost}

\begin{figure*}[t]
\centering
\includegraphics[width=\textwidth]{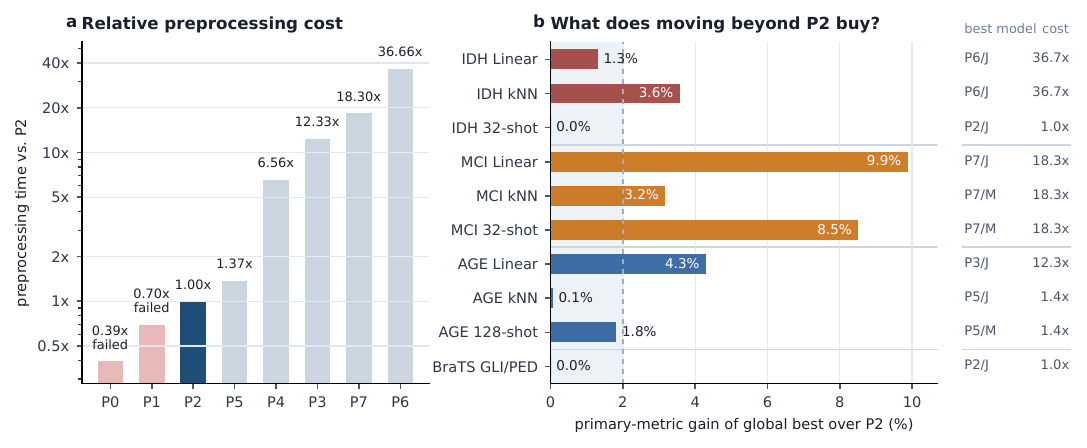}
\caption{P2-centered cost--utility analysis. (a) Relative preprocessing time for each P-level, normalized by P2. P0 and P1 failed with numerical instability and loss collapse, making P2 the lowest-cost feasible level. (b) Primary-metric gain of the global-best P-level over P2 for representative downstream settings, grouped by task. Most settings show small gains from heavier preprocessing, whereas MCI exhibits the largest gains but requires the substantially more expensive P7 pipeline.}
\label{fig:p2-cost-utility}
\end{figure*}

We next ask the central cost--utility question: after P2 has been established as the lowest-cost feasible preprocessing level, how much downstream performance is gained by moving to heavier pipelines? Figure~\ref{fig:p2-cost-utility} summarizes this P2-centered comparison. Panel (a) reports preprocessing time normalized by P2. Although P0 and P1 are cheaper, they failed under the current standardized pretraining setup because of NaN gradients and loss collapse; P2 is therefore the first viable point in the preprocessing spectrum. The qualitative changes in pretraining inputs across the spectrum are shown in the supplementary material. Beyond P2, cost rises sharply: P5 costs 1.37$\times$ P2, P4 costs 6.56$\times$, P3 costs 12.33$\times$, P7 costs 18.30$\times$, and P6 costs 36.66$\times$.

Panel (b) shows the primary-metric gain obtained by replacing the best P2 model with the global-best P-level model in each representative downstream setting. The dominant pattern is that most settings gain little from moving beyond P2. IDH linear probing improves by 1.3\%, IDH kNN by 3.6\%, and IDH 32-shot by 0.0\%. AGE is also close to saturation at P2 in most settings: AGE kNN improves by only 0.1\%, AGE 128-shot by 1.8\%, and BraTS GLI/PED macro Dice shows no gain. AGE linear regression is a moderate exception, with a 4.3\% MAE reduction, but the best result requires P3 and therefore a 12.3$\times$ preprocessing cost.

MCI is the clearest setting in which heavier preprocessing provides meaningful utility. MCI linear probing gains 9.9\%, and MCI 32-shot gains 8.5\%; both gains are achieved by P7 models, requiring approximately 18.3$\times$ the preprocessing cost of P2. Thus, the cost--utility result is not that preprocessing should always be minimized. Rather, escalation beyond P2 should be justified by task-specific gains large enough to offset the added preprocessing cost. P2 serves as a low-cost feasible anchor that preserves most performance in many settings, while heavier preprocessing is warranted only when the downstream gain is substantial.

As an exploratory paired uncertainty analysis, we further compared P2 with the empirically best P-level in each primary setting. Table~\ref{tab:p2-vs-best-uncertainty-main} reports representative comparisons, with the full set in the supplementary material. Across 24 P2-vs-best comparisons, only 7 had 95\% confidence intervals excluding zero, only one reached nominal significance, and only one survived BH correction. Many apparent gains over P2 are therefore statistically unresolved and practically marginal relative to the additional preprocessing cost.

\begin{table*}[!htbp]
\centering
\small
\setlength{\tabcolsep}{3pt}
\begin{tabular}{@{}lllllll@{}}
\toprule
Setting & Comparison & P2 & Best & Improvement 95\% CI & p & BH q \\
\midrule
IDH LINEAR JEPA & P2 vs P6 & 0.8652 & \textbf{0.8767} & +0.0115 [-0.0238, +0.0496] & 0.6875 & 0.9706 \\
MCI LINEAR JEPA & P2 vs P7 & 0.6756 & \textbf{0.7533} & +0.0778 [-0.0370, +0.2093] & 0.3750 & 0.6923 \\
AGE LINEAR JEPA & P2 vs P3 & 2.9800 & \textbf{2.7596} & +0.2173 [+0.1402, +0.2928] & $<0.0001$ & 0.0012 \\
AGE KNN JEPA & P2 vs P5 & 2.4691 & \textbf{2.4676} & +0.0018 [-0.0379, +0.0422] & 0.9295 & 1.0000 \\
BRATS GLI+PED macro MAE & P2 vs P5 & 0.7756 & \textbf{0.7776} & +0.0020 [-0.0036, +0.0083] & 0.5442 & 0.8708 \\
BRATS GLI+PED macro JEPA & P2 vs P2 & \textbf{0.8033} & \textbf{0.8033} & +0.0000 [+0.0000, +0.0000] & 1.0000 & 1.0000 \\
\bottomrule
\end{tabular}
\caption{Selected P2-vs-best paired uncertainty results. Positive improvement means that the empirically best P-level improves over P2; for AGE MAE, positive values indicate error reduction.}
\label{tab:p2-vs-best-uncertainty-main}

\end{table*}

\subsection{MCI Shows the Clearest Task-Specific Exception}

The residual benefit of heavier preprocessing is strongly task-dependent. As shown in Figure~\ref{fig:p2-cost-utility}b, IDH shows modest gains from higher P-levels, AGE is often nearly saturated at P2, and GLI/PED segmentation is already maximized by P2 under the fixed-input transfer setting. MCI is the clearest empirical exception: its largest observed improvements come from P7, which combines N4 correction, brain extraction, and affine alignment. This pattern raises a more precise question than whether P7 is better for MCI: does the MCI gain require high-cost P7 preprocessing throughout foundation-model pretraining, or can it be recovered primarily through stronger downstream input standardization?

To separate pretraining-side and downstream-side effects, we performed a cross-level transfer experiment pairing P2- and P7-pretrained checkpoints with P2- and P7-processed MCI downstream data. Table~\ref{tab:mci-cross-level-main} reports linear-probe AUROC for both JEPA and MAE, with additional metrics in the supplementary material. For JEPA, the P2 checkpoint improves from 0.6756 AUROC on P2 MCI data to 0.7289 on P7 MCI data, closing 68.6\% of the gap to the matched P7-checkpoint/P7-data reference value of 0.7533. The reverse transfer is not robust: the P7 checkpoint drops to 0.6378 AUROC when evaluated on P2 data, below the P2-checkpoint/P2-data reference. MAE shows the same pattern. The P2 checkpoint improves from 0.6789 to 0.7330 AUROC when downstream data are processed at P7, closing 81.1\% of the matched-reference gap, whereas the P7 checkpoint drops to 0.6574 AUROC on P2 data. An additional overlap-controlled analysis removes all pretraining volumes that are also present in the MCI evaluation set, and the same cross-level transfer pattern persists; full results are reported in the supplementary material.

\begin{table}[!htbp]
\centering
\small
\setlength{\tabcolsep}{1mm}
\begin{tabular}{@{}ll@{\hspace{1mm}}cc@{\hspace{1mm}}cc@{}}
\toprule
Objective & Checkpoint & P2 data & P7 data & Off-diag. $\Delta$ & Gap closure \\
\midrule
JEPA & P2 ckpt & \textbf{0.6756} & 0.7289 & +0.0533 & 68.6\% \\
JEPA & P7 ckpt & 0.6378 & \textbf{0.7533} & -0.1156 & -- \\
\addlinespace[1pt]
MAE & P2 ckpt & \textbf{0.6789} & 0.7330 & +0.0541 & 81.1\% \\
MAE & P7 ckpt & 0.6574 & \textbf{0.7456} & -0.0881 & -- \\
\bottomrule
\end{tabular}
\caption{MCI cross-level transfer under linear probing. Entries are AUROC. Off-diagonal $\Delta$ is computed relative to the same-checkpoint matched-reference cell: P2 checkpoint on P7 data versus P2 checkpoint on P2 data, or P7 checkpoint on P2 data versus P7 checkpoint on P7 data. Gap closure is reported for the P2 checkpoint.}
\label{tab:mci-cross-level-main}

\end{table}

These findings refine the interpretation of MCI as the clearest observed exception to the P2-centered cost--utility pattern. The exception is task-specific and partly downstream-side: MCI appears to benefit from stronger spatial standardization, but this does not imply that the entire foundation pretraining pipeline must use P7. A more cost-effective design is to use P2 as the default large-scale pretraining level and reserve P7-like processing for downstream tasks, such as MCI, whose signal appears particularly sensitive to anatomical alignment and brain-focused input standardization.

\subsection{Self-Supervised Objectives Modulate Preprocessing Sensitivity}

\begin{figure*}[t]
\centering
\includegraphics[width=0.82\textwidth]{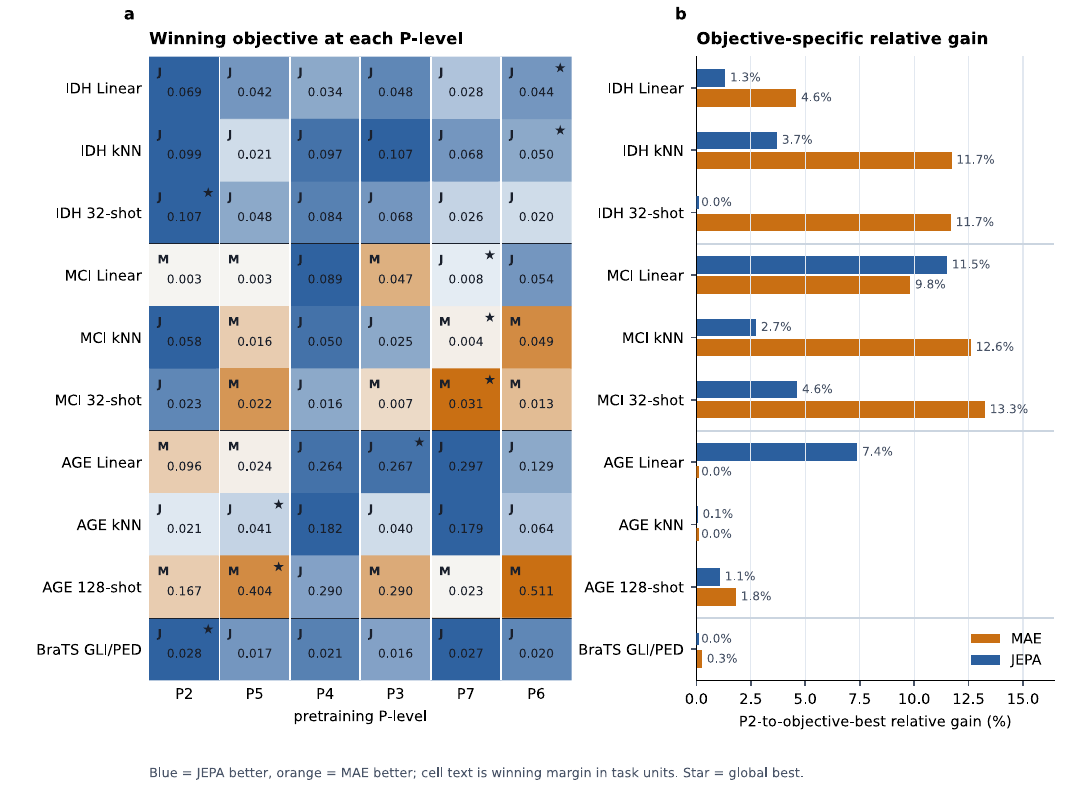}
\caption{Objective interaction across preprocessing levels. (a) Winning objective at each P-level for representative downstream settings. (b) Objective-specific relative gain from P2 to the best P-level within the same objective.}
\label{fig:objective-interaction}
\end{figure*}

The preceding analyses show that heavier preprocessing is useful only in selected downstream settings. A remaining question is whether this behavior is a property of the input pipeline alone, or whether it also depends on the objective used to learn the representation. This distinction matters because MAE and JEPA impose different learning pressures: MAE reconstructs masked image content and may therefore remain more exposed to local intensity, texture, and bias-field variation, whereas JEPA predicts latent targets and may emphasize higher-level anatomical organization. We therefore analyze preprocessing effects separately for each objective in Figure~\ref{fig:objective-interaction}.

The clearest objective-level pattern is that JEPA provides strong endpoints in settings where P2 is already a good preprocessing anchor. In IDH, JEPA is better than MAE at P2 across linear probing, kNN, and 32-shot evaluation, and the best IDH linear and kNN results remain JEPA P6 (0.8767 and 0.8511 AUROC). In fixed-input GLI/PED segmentation, JEPA P2 is itself the global best, reaching 0.8033 macro Dice, and heavier preprocessing does not improve the JEPA segmentation result. These results suggest that, once basic spatial and intensity standardization is available, JEPA can often learn transferable representations without relying on increasingly aggressive preprocessing. At the same time, JEPA's stronger endpoint should not be confused with greater preprocessing gain: in IDH, MAE improves more than JEPA when allowed to move beyond P2, but the improved MAE models still do not overtake JEPA.

MCI and AGE show why objective choice cannot be treated as a secondary detail. In MCI kNN and 32-shot evaluation, JEPA is better at P2, but the best endpoints after preprocessing selection are MAE P7, reaching 0.7178 and 0.7205 AUROC. This means that the MCI exception is not only a task effect; it is also shaped by how the representation was learned. AGE shows a different form of coupling. MAE is already optimal at P2 for AGE linear and kNN regression, whereas JEPA has a sizable gain in AGE linear when moving from P2 to P3, reducing MAE from 2.9800 to 2.7596. In AGE few-shot 128, the best endpoint is instead MAE P5 with prediction error 3.4162. Thus, heavier preprocessing is not uniformly ``MAE-favoring'' or ``JEPA-favoring''; its effect changes with both the downstream protocol and the pretraining objective.

This objective dependence sharpens the main cost--utility conclusion. P2 remains a strong low-cost anchor, but the reason to leave P2 is not determined by task identity alone. It also depends on whether the encoder was trained with MAE or JEPA, and on the downstream protocol used to read out the representation. Preprocessing should therefore be selected jointly with the self-supervised objective rather than treated as an objective-independent input choice.

\section{Discussion}

The central implication of this study is that MRI preprocessing should be treated as a costed modeling choice, not as a fixed prerequisite for brain foundation models. In our controlled sweep, P2 is the first preprocessing level that provides stable large-scale pretraining, while P0/P1 fail under the current training assumptions. This does not mean that raw or weakly normalized MRI can never be made trainable; rather, it shows that some minimum standardization of geometry and foreground intensity is practically important for the tested 3D ViT, MAE, and JEPA setup. Once this minimum is reached, however, escalating preprocessing is not automatically rewarded. The aggregate gain from replacing P2 with the best feasible level is small, and most paired improvements over P2 remain statistically unresolved relative to the additional preprocessing cost.

The results also show why a single preprocessing prescription is unlikely to fit all downstream uses. MCI is the clearest empirical case where stronger anatomical standardization helps, but the cross-level transfer experiment changes the interpretation of that gain: much of the P7 advantage can be recovered by applying P7-like processing to the downstream MCI data while retaining P2 pretraining. By contrast, AGE and GLI/PED fixed-input segmentation are often already near or best at P2, suggesting that heavier spatial normalization may offer little benefit when the task depends on individual variation or local spatial structure. IDH occupies an intermediate regime, with modest gains from higher P-levels but not enough evidence to justify a universal move away from P2. The interaction between MAE and JEPA further reinforces this point: preprocessing effects are coupled to the representation objective rather than being a property of the input pipeline alone.

These findings have methodological implications for the evaluation of brain MRI foundation models. Preprocessing should be reported as part of model design, because it defines the input distribution, computational cost, numerical stability, and the biological or acquisition-related signals preserved in the input. A P2-centered analysis provides one practical way to compare task-specific utility against the cost of stronger standardization.

\section{Limitations and Future Work}

This study is limited by the scope of the pretraining and evaluation design. P0/P1 are infeasible in our current implementation because they lead to NaN gradients and loss collapse, but this may depend on patchification, intensity scaling, optimizer settings, or loss normalization. Future work should test whether weaker preprocessing can be stabilized by alternative input normalization or training objectives, thereby separating implementation constraints from fundamental limits. We also use one 3D ViT capacity, two self-supervised objectives, and a 20,000-volume sample from FOMO300K \citep{fomomri2026fomo300k}; validating the same conclusions with larger encoders, larger samples, additional objectives, and multimodal or longitudinal settings remains necessary. Future studies should also test the same cost--utility pattern on additional fully external downstream cohorts.

The downstream evidence is also intentionally bounded. MCI has the smallest case-level sample size, so its P7 advantage should be confirmed in larger and external cohorts. BraTS is a fixed-input transfer setting because the released data are already substantially standardized, and preprocessing cost is measured for our implementation and hardware.

\section{Conclusion}

MRI preprocessing changes the input distribution, the representation learned during self-supervised pretraining, and the cost of building brain foundation models. In a controlled P0--P7 sweep, we find that P2 is the lowest-cost feasible level and often preserves most of the best observed downstream performance. More aggressive preprocessing is useful only in selected regimes, most clearly MCI, and even there much of the gain can be recovered at downstream preprocessing time. The practical conclusion is therefore not to minimize preprocessing or to maximize it by default, but to choose preprocessing strength as a downstream-aware cost--utility decision.

\begingroup
\small
\bibliographystyle{unsrtnat}
\bibliography{references}
\endgroup

\clearpage
\appendix
\raggedbottom
\setcounter{figure}{0}
\setcounter{table}{0}
\renewcommand{\thefigure}{A\arabic{figure}}
\renewcommand{\thetable}{A\arabic{table}}
\small
\setlength{\textfloatsep}{6pt plus 1pt minus 2pt}
\setlength{\floatsep}{6pt plus 1pt minus 2pt}
\setlength{\intextsep}{6pt plus 1pt minus 2pt}
\setlength{\dbltextfloatsep}{6pt plus 1pt minus 2pt}
\setlength{\dblfloatsep}{6pt plus 1pt minus 2pt}
\captionsetup{font=footnotesize,skip=2pt}

\section{Appendix: Additional Results and Details}

\subsection{Pretraining Implementation Details}

Table~\ref{tab:pretraining-corpus-composition} reports the source-level composition of the 20,000-volume pretraining corpus. Table~\ref{tab:preprocess} gives the concrete P0--P7 image definitions and measured preprocessing time used by the pretraining sweep. P0--P2 are applied from raw records during training, whereas P3--P7 are generated as offline manifests and then loaded as already-preprocessed volumes. All levels are cropped or padded to the same fixed input size before patchification. Table~\ref{tab:pretraining-failure-diagnostics} summarizes the P0/P1 stability diagnostics used to define the feasible boundary.

\begin{table}[!htbp]
\centering
\scriptsize
\setlength{\tabcolsep}{3pt}
\renewcommand{\arraystretch}{0.96}
\begin{tabular}{@{}L{0.30\textwidth}rrL{0.30\textwidth}rr@{}}
\toprule
Dataset & Samples & Share & Dataset & Samples & Share \\
\midrule
\texttt{PT030\_OpenNeuro} & 12,335 & 61.68\% & \texttt{PT021\_IXI} & 187 & 0.94\% \\
\texttt{PT018\_HBN} & 2,222 & 11.11\% & \texttt{PT005\_ADHD\_200} & 157 & 0.78\% \\
\texttt{PT009\_BraTS-GEN} & 1,027 & 5.13\% & \texttt{PT002\_Nigerian\_Clinical} & 108 & 0.54\% \\
\texttt{PT027\_NKI} & 786 & 3.93\% & \texttt{PT013\_Calgary\_Preschool} & 90 & 0.45\% \\
\texttt{PT020\_HCP\_Wu\_Minn} & 555 & 2.77\% & \texttt{PT008\_AdolescentBrainDevelopment} & 49 & 0.24\% \\
\texttt{PT014\_CoRR} & 414 & 2.07\% & \texttt{PT012\_CHBMP} & 33 & 0.17\% \\
\texttt{PT029\_OASIS2} & 341 & 1.70\% & \texttt{PT017\_HBN-SSI} & 31 & 0.15\% \\
\texttt{PT024\_M4Raw} & 301 & 1.50\% & \texttt{PT022\_Bejing\_Enhanced} & 29 & 0.14\% \\
\texttt{PT016\_GSP} & 275 & 1.38\% & \texttt{PT019\_HCP\_Wu\_Minn\_Test\_Retest} & 23 & 0.11\% \\
\texttt{PT023\_Infant\_Development\_Brain} & 268 & 1.34\% & \texttt{PT004\_ACPI} & 20 & 0.10\% \\
\texttt{PT028\_OASIS1} & 264 & 1.32\% & \texttt{PT026\_MICA\_MICs} & 16 & 0.08\% \\
\texttt{PT015\_MSD\_BrainTumor} & 259 & 1.29\% & \texttt{PT001\_ClevelandCCF} & 5 & 0.03\% \\
\texttt{PT010\_BrainLat} & 203 & 1.01\% & \texttt{PT003\_CUNMET} & 2 & 0.01\% \\
\bottomrule
\end{tabular}
\caption{Source-level composition of the 20,000-volume self-supervised pretraining corpus sampled from the gated FOMO300K superset. The table lists all constituent source datasets with nonzero sampled volumes.}
\label{tab:pretraining-corpus-composition}

\end{table}

\begin{table}[!htbp]
\centering
\small
\begin{tabular}{@{}lL{0.46\textwidth}rr@{}}
\hline
Level & Main operation & Seconds/sample & Cost vs. P2 \\
\hline
P0 & Foreground ROI crop/pad only; native acquisition properties retained & 1.92 & 0.39$\times$ \\
P1 & P0 + basic spatial standardization & 3.40 & 0.70$\times$ \\
P2 & P1 + foreground percentile clipping and z-score normalization & 4.87 & 1.00$\times$ \\
P3 & P2 + N4 bias-field correction & 60.07 & 12.33$\times$ \\
P4 & P2 + brain extraction & 31.94 & 6.56$\times$ \\
P5 & P2 + affine template alignment & 6.69 & 1.37$\times$ \\
P6 & P2 + affine-initialized B-spline deformable alignment & 178.54 & 36.66$\times$ \\
P7 & P2 + N4, brain extraction, and affine alignment & 89.12 & 18.30$\times$ \\
\hline
\end{tabular}
\caption{Preprocessing levels and measured mean time per MRI volume. P0/P1 are cheaper but are treated as infeasible because raw, unnormalized intensity distributions caused numerical instability, NaN gradients, and loss collapse under the standardized pretraining setup.}
\label{tab:preprocess}

\end{table}

\begin{table}[!htbp]
\centering
\footnotesize
\setlength{\tabcolsep}{3pt}
\begin{tabular}{@{}lL{0.21\textwidth}L{0.21\textwidth}L{0.34\textwidth}@{}}
\toprule
Level & MAE status & JEPA status & Diagnostic interpretation \\
\midrule
P0 & Failed before completing one epoch & Failed before completing one epoch & Raw native intensity scale led to NaN gradients and loss collapse under the fixed pretraining recipe. \\
P1 & Failed before completing one epoch & Failed before completing one epoch & Basic spatial standardization alone did not stabilize optimization without foreground intensity clipping and normalization. \\
P2 & Stable for 100 epochs & Stable for 100 epochs & Adding foreground percentile clipping and z-score normalization made both objectives train stably with the same backbone, masking protocol, precision, and optimizer family. \\
\bottomrule
\end{tabular}
\caption{Pretraining stability diagnostics for the P0--P2 boundary. P0/P1 were not excluded because of weak downstream performance; they failed the fixed self-supervised pretraining protocol itself. P2 is therefore treated as the first feasible preprocessing level in this study.}
\label{tab:pretraining-failure-diagnostics}

\end{table}

All self-supervised runs use the same single-channel 3D ViT-Base encoder and foreground-aware contiguous block masking; only the preprocessing level and self-supervised objective vary. The implementation passes a patch-validity mask to the encoder so that global feature pooling and attention ignore patches whose foreground fraction is below 0.05. Table~\ref{tab:app-pretraining-hparams} summarizes the objective-specific settings used across P-levels.

\begin{table}[!htbp]
\centering
\footnotesize
\setlength{\tabcolsep}{4pt}
\begin{tabular}{L{0.22\textwidth}L{0.35\textwidth}L{0.35\textwidth}}
\toprule
Setting & MAE & JEPA \\
\midrule
Encoder & 3D ViT-Base; 12 layers; hidden size 768; 12 attention heads & 3D ViT-Base; 12 layers; hidden size 768; 12 attention heads \\
Input and patches & $192^3$ input; non-overlapping $16^3$ patches; foreground threshold 0.05 & $192^3$ input; non-overlapping $16^3$ patches; foreground threshold 0.05 \\
Masking & Mask ratio 0.5; four contiguous 3D mask blocks; at least four new patch tokens per block & Mask ratio 0.5; four contiguous 3D mask blocks; at least four new patch tokens per block \\
Prediction target & Voxel reconstruction for masked patches & Latent target representation for masked patches \\
Loss & MSE reconstruction loss with normalized patch targets & L1 latent prediction loss without target L2 normalization \\
Objective head & Decoder hidden size 512; 8 layers; 16 heads & Predictor hidden size 384; 6 layers; 6 heads \\
Teacher network & None & EMA teacher; fixed momentum 0.99925 \\
Training length & 100 epochs & 100 epochs \\
Batching & Per-device batch size 10; gradient accumulation 1; 4 dataloader workers & Per-device batch size 10; gradient accumulation 1; 4 dataloader workers \\
Precision and seed & bf16 for feasible P-levels; seed 42 & bf16; seed 239 \\
Optimizer & AdamW; learning rate $2\times10^{-5}$; cosine schedule; 10240 warmup steps & AdamW; peak learning rate $6\times10^{-4}$; start learning rate $1\times10^{-4}$; cosine schedule; 40 warmup epochs \\
Regularization & Weight decay 0.05; gradient clipping max norm 1.0 & Weight decay 0.04; gradient clipping max norm 1.0 \\
\bottomrule
\end{tabular}
\caption{Self-supervised pretraining hyperparameters used for the PreBrain P-level sweep. Within each objective, the settings are fixed across preprocessing levels so that P-level comparisons isolate input standardization rather than model capacity or optimization changes.}
\label{tab:app-pretraining-hparams}

\end{table}

Hardware and software environment.
All self-supervised pretraining jobs were launched with eight NVIDIA A100-PCIE-40GB GPUs. Each inspected execution node exposed four NVIDIA A100-PCIE-40GB devices, each with 40960 MiB of device memory; PyTorch reported approximately 39.39 GiB per device. The CPU platform was a dual-socket HiSilicon Kunpeng-920 system with $2\times64$ cores, for 128 CPU cores in total, using the aarch64 architecture. The operating system was Kylin Linux Advanced Server V10 (Sword), with kernel 4.19.90-24.4.v2101.ky10.aarch64. The NVIDIA driver version was 535.104.12, \texttt{nvidia-smi} reported CUDA 12.2, PyTorch was built against CUDA 12.9, and cuDNN was version 9.10.2.21, reported by PyTorch as 91002. The Python environment used Python 3.12.0, pip 25.3, PyTorch 2.7.1, torchvision 0.22.1, torchaudio 2.8.0, MONAI 1.5.2, nibabel 5.4.2, SimpleITK 2.5.3, SciPy 1.17.1, pandas 2.3.3, and scikit-learn 1.8.0.

\FloatBarrier

\subsection{Downstream Protocol Details}

Downstream preprocessing follows the input definition used by each evaluated model during training or pretraining. Thus pretrained MAE/JEPA checkpoints are evaluated with the same preprocessing definition used to produce their pretraining data; ViT-Scratch controls are randomly initialized and trained end-to-end under the same input definition within each run; and baseline models use the input protocol associated with their own released or training configuration. This keeps comparisons input-consistent within each model family. BraTS uses the released image/mask geometry without reorientation or spacing resampling; pretrained segmentation transfer freezes the encoder, whereas the ViT-Scratch segmentation control trains the encoder and decoder jointly.

For IDH, MCI, and AGE, frozen evaluation first extracts one global feature vector per case. PreBrain encodes each modality independently; multi-modal IDH features are concatenated before the downstream predictor. The same train/validation/test fold assignment is used across evaluated models. Validation data are used for hyperparameter selection, early stopping, and classification-threshold selection; test folds are held out until final metric computation. Table~\ref{tab:overlap-audit} reports the exact file-path overlap audit between the self-supervised pretraining corpus and downstream evaluations.

\begin{table}[!htbp]
\centering
\scriptsize
\setlength{\tabcolsep}{2.8pt}
\begin{tabular}{L{0.15\textwidth}L{0.21\textwidth}L{0.15\textwidth}L{0.27\textwidth}L{0.16\textwidth}}
\toprule
Task & Data and label & Input & Split and input consistency & Reported metrics \\
\midrule
IDH classification & 495 UCSF-PDGM baseline cases; IDH-wildtype is negative and all other non-empty IDH labels are positive & FLAIR and T1c encoded separately, then fused at case level & Five outer folds; validation is about 10\% of the development pool; stratified by IDH label and WHO CNS grade. Downstream inputs follow the training or pretraining input definition of the evaluated model & AUROC and AUPRC, with accuracy, balanced accuracy, F1, sensitivity, and specificity \\
MCI classification & 235 OASIS-1 T1/T1w samples with binary cognitive-impairment labels & Single T1/T1w volume & Five outer folds; validation is about 10\% of the development pool; stratified by label. Downstream inputs follow the training or pretraining input definition of the evaluated model & AUROC and AUPRC, with accuracy, balanced accuracy, F1, sensitivity, and specificity \\
AGE regression & 3578 T1/T1w samples from HCP S1200, LONG579, SALD, Calgary, PETfrog, and IXI \citep{hcpS1200,wang2022longitudinal,wei2018sald,reynolds2020calgary,luna2020petfrog,ixiDataset}; ages 2--39 years after age filtering & Single T1/T1w volume & Five subject-grouped outer folds; validation is about 10\% of the development pool; stratified by dataset and 5-year age bin. Downstream inputs follow the training or pretraining input definition of the evaluated model & MAE and Pearson correlation, with RMSE, MSE, $R^2$, and mean error \\
BraTS segmentation & BraTS-GLI (1251 cases) \citep{baid2021brats} and BraTS-PED (99 cases) \citep{bratspedsTcia} T2-FLAIR images with binary tumor masks & T2-FLAIR image; dense encoder feature maps for decoder training & Dataset-specific train/validation/test splits with fractions 0.7/0.1/0.2 and seed 42. GLI uses 876/125/250 cases and PED uses 69/10/20 cases. Inputs keep the released geometry and use Otsu foreground estimation plus P2-style intensity-only normalization & Dice and IoU on held-out test cases \\
\bottomrule
\end{tabular}
\caption{Downstream task definitions and input-consistency protocol. Downstream preprocessing follows the training or pretraining input definition of the evaluated model.}
\label{tab:app-downstream-protocols}

\end{table}

\begin{table}[!htbp]
\centering
\footnotesize
\setlength{\tabcolsep}{4pt}
\begin{tabular}{@{}L{0.17\textwidth}L{0.34\textwidth}L{0.18\textwidth}L{0.23\textwidth}@{}}
\toprule
Task & Overlap source in pretraining corpus & Exact overlap & Note \\
\midrule
MCI & \texttt{PT028\_OASIS1} & 37 / 235 (15.7\%) & Partial exact overlap \\
IDH & None & 0 / 495 (0.0\%) & No exact overlap \\
AGE & \texttt{PT020\_HCP\_Wu\_Minn}, \texttt{PT021\_IXI} & 312 / 3,578 (8.7\%) & Partial exact overlap \\
BraTS GLI/PED & None & 0 / 1,350 (0.0\%) & No exact overlap \\
\bottomrule
\end{tabular}
\caption{Exact file-path overlap between the self-supervised pretraining corpus and downstream evaluations. Exact overlap denotes downstream image files that also appear in the pretraining corpus; downstream labels, fold assignments, validation metrics, and test outcomes are not used during self-supervised pretraining.}
\label{tab:overlap-audit}

\end{table}

\begin{table}[!htbp]
\centering
\scriptsize
\setlength{\tabcolsep}{3pt}
\begin{tabular}{L{0.22\textwidth}L{0.22\textwidth}L{0.50\textwidth}}
\toprule
Evaluation mode & Tasks & Hyperparameters and selection rule \\
\midrule
kNN frozen probe & IDH, MCI, AGE & L2-normalized features; $k\in\{1,5,10,20,50\}$; temperature 0.07. Classification uses class-balanced voting and selects $k$ by validation AUROC. AGE uses similarity-weighted averaging and selects $k$ by validation MAE. \\
Linear frozen probe & IDH, MCI, AGE & Features are standardized using training-set statistics. Classification uses a logistic head with weighted BCE; AGE uses a linear regressor with target standardization and MSE. AdamW learning rate is $10^{-4}$; weight decay is selected from $\{0,10^{-4},10^{-3},10^{-2},10^{-1}\}$; maximum 300 epochs; patience 40. \\
Few-shot frozen probe & IDH, MCI, AGE & Uses the same linear-probe training recipe after selecting weight decay on the full training split. Classification samples 1, 2, 4, 8, 16, or 32 examples per class; AGE samples 8, 16, 32, 64, or 128 total examples. Each shot is repeated 10 times with fixed seeds. \\
From-scratch full fine-tuning control & IDH, MCI, AGE, BraTS-GLI, BraTS-PED & ViT-Scratch is randomly initialized for all tasks. For IDH, MCI, and AGE, batch size is 2 for 40 epochs; during the first 5 warmup epochs the encoder is frozen and only the task head is trained, after which encoder and head are optimized jointly. Backbone learning rate is $10^{-4}$; head learning rate is $10^{-3}$; weight decay is $10^{-4}$; dropout is 0.2; gradient clipping is 1.0; early-stopping patience is 8, using validation AUROC for IDH/MCI and validation MAE for AGE. For BraTS, the scratch encoder and dense decoder are trained jointly with \texttt{freeze\_encoder=false}; patch size is $128^3$; positive-crop probability is 0.7; batch size is 1; training lasts 20 epochs with learning rate $10^{-4}$, weight decay $10^{-4}$, decoder channels 128, Dice+BCE loss, gradient clipping 1.0, sliding-window overlap 0.5, and threshold 0.5. \\
Frozen dense decoder transfer & BraTS-GLI, BraTS-PED & For pretrained encoder transfer, the encoder is frozen with \texttt{freeze\_encoder=true} and a lightweight 3D decoder is trained on dense feature maps. Patch size $128^3$; positive-crop probability 0.7; batch size 1; AdamW learning rate $10^{-4}$; weight decay $10^{-4}$; 20 epochs; decoder channels 128; Dice+BCE loss; gradient clipping 1.0; sliding-window inference with overlap 0.5; threshold 0.5. \\
\bottomrule
\end{tabular}
\caption{Downstream optimization details. All validation choices are made inside each training fold before test evaluation, and inputs are kept consistent with the evaluated model's training or pretraining definition.}
\label{tab:app-downstream-optimization}

\end{table}

\FloatBarrier

\subsection{Pretraining Input Visualization}

Figure~\ref{fig:app-pretraining-inputs} shows representative input volumes after each preprocessing level. It is intended as a qualitative companion to Table~\ref{tab:preprocess}: P2 is the first level that standardizes both geometry and foreground intensity, whereas P3--P7 add bias correction, brain extraction, affine alignment, deformable alignment, or combined aggressive standardization.

\begin{center}
\includegraphics[width=0.94\textwidth]{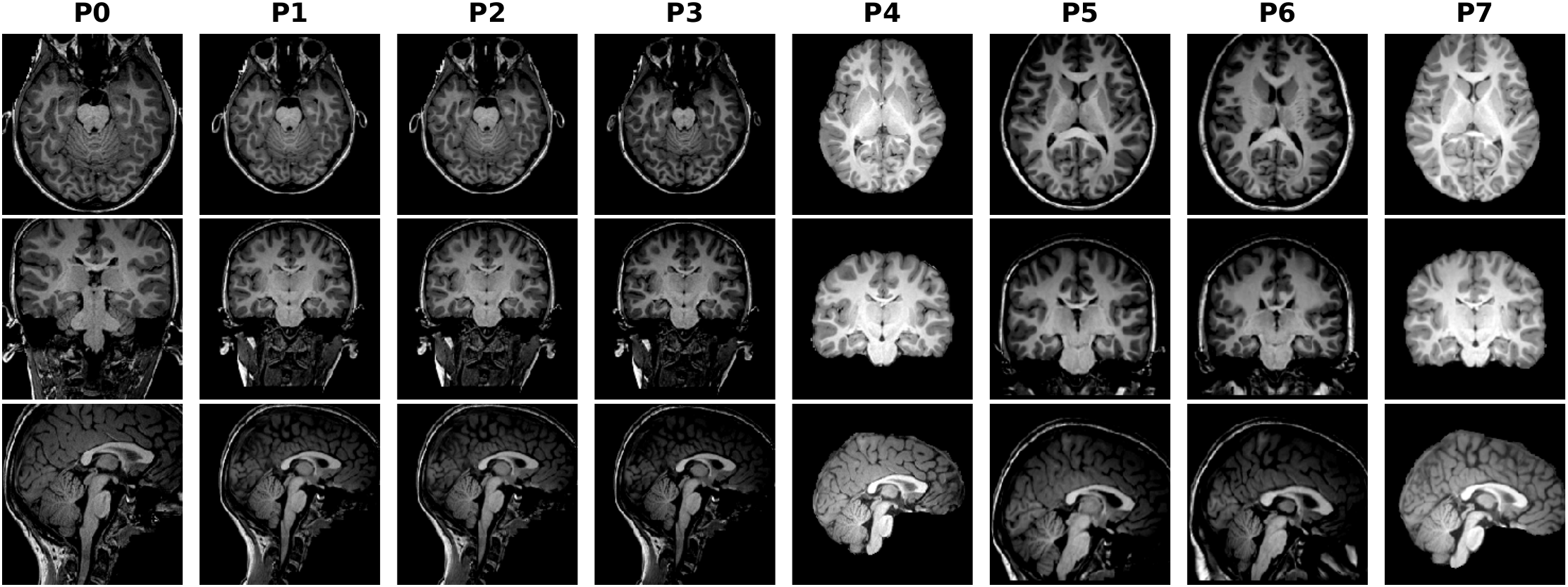}
\captionof{figure}{Representative pretraining inputs across the P0--P7 preprocessing spectrum. The visualization illustrates how increasing preprocessing strength changes field of view, intensity scaling, non-brain tissue content, and anatomical alignment before self-supervised pretraining.}
\label{fig:app-pretraining-inputs}
\end{center}

\FloatBarrier

\subsection{Baseline Details}

Tables~\ref{tab:app-idh-baselines}--\ref{tab:app-brats-gli-ped-baselines} provide the full baseline details supporting the baseline sanity check in the main paper. For IDH and MCI, the pretrained row in each setting is the best MAE/JEPA checkpoint selected by AUROC. For AGE, the pretrained row is selected by MAE. Baseline checkpoints use the same BrainIAC, GenBrain, MedicalNet/Med3D, MRI-Core, and RadFM models described in the main text, with downstream inputs prepared according to each model's own training or released configuration. The scratch row reports ViT-Scratch trained from random initialization on the downstream labels under the same folds and model-consistent input definition used for that run; for BraTS, the scratch encoder and dense decoder are trained jointly with \texttt{freeze\_encoder=false}. BraTS is reported as fixed-input segmentation on GLI and PED only.

\begin{table}[!htbp]
\centering
\scriptsize
\setlength{\tabcolsep}{2.2pt}
\renewcommand{\arraystretch}{0.95}
\begin{tabular}{@{}lllrrrrrrr@{}}
\toprule
Setting & Type & Model & AUROC & AUPRC & Acc. & BAcc. & F1 & Sens. & Spec. \\
\midrule
Linear & Ours & JEPA P6 & \textbf{0.8767} & \textbf{0.6941} & \textbf{0.8433} & 0.7728 & \textbf{0.6442} & 0.6523 & \textbf{0.8933} \\
 & Ext. & BrainiAC & 0.7486 & 0.4673 & 0.7257 & 0.6712 & 0.4631 & 0.5794 & 0.7630 \\
 & Ext. & GenBrain & 0.8186 & 0.6015 & 0.7492 & 0.7219 & 0.5388 & 0.6759 & 0.7679 \\
 & Ext. & MedicalNet & 0.8332 & 0.6442 & 0.7890 & \textbf{0.7788} & 0.6009 & \textbf{0.7612} & 0.7965 \\
 & Ext. & MRI-Core & 0.8547 & 0.6689 & 0.8185 & 0.7549 & 0.5975 & 0.6451 & 0.8648 \\
 & Ext. & RadFM & 0.7586 & 0.4663 & 0.7076 & 0.6596 & 0.4455 & 0.5769 & 0.7424 \\
\addlinespace[1pt]
kNN & Ours & JEPA P6 & \textbf{0.8511} & \textbf{0.6536} & \textbf{0.7885} & \textbf{0.7542} & \textbf{0.5825} & 0.6943 & \textbf{0.8140} \\
 & Ext. & BrainiAC & 0.7037 & 0.4286 & 0.6753 & 0.6513 & 0.4382 & 0.6108 & 0.6918 \\
 & Ext. & GenBrain & 0.5767 & 0.2955 & 0.5868 & 0.5318 & 0.2949 & 0.4405 & 0.6232 \\
 & Ext. & MedicalNet & 0.6096 & 0.3372 & 0.6531 & 0.5832 & 0.3489 & 0.4646 & 0.7018 \\
 & Ext. & MRI-Core & 0.7924 & 0.5809 & 0.7577 & 0.7353 & 0.5584 & \textbf{0.6952} & 0.7753 \\
 & Ext. & RadFM & 0.6647 & 0.3480 & 0.6526 & 0.6061 & 0.3787 & 0.5259 & 0.6864 \\
\addlinespace[1pt]
1-shot & Ours & MAE P6 & \textbf{0.5361} & \textbf{0.2434} & 0.5135 & \textbf{0.5374} & \textbf{0.3003} & 0.5771 & 0.4977 \\
 & Ext. & BrainiAC & 0.5054 & 0.2203 & 0.4774 & 0.5080 & 0.2754 & 0.5603 & 0.4556 \\
 & Ext. & GenBrain & 0.5113 & 0.2246 & 0.4822 & 0.5161 & 0.2873 & 0.5738 & 0.4584 \\
 & Ext. & MedicalNet & 0.5016 & 0.2147 & 0.4609 & 0.5067 & 0.2785 & \textbf{0.5830} & 0.4304 \\
 & Ext. & MRI-Core & 0.5040 & 0.2279 & 0.5261 & 0.5189 & 0.2739 & 0.5047 & 0.5331 \\
 & Ext. & RadFM & 0.5069 & 0.2257 & \textbf{0.5308} & 0.5069 & 0.2480 & 0.4663 & \textbf{0.5476} \\
\addlinespace[1pt]
2-shot & Ours & JEPA P2 & \textbf{0.6439} & \textbf{0.3717} & 0.6102 & \textbf{0.5949} & \textbf{0.3724} & \textbf{0.5692} & 0.6206 \\
 & Ext. & BrainiAC & 0.5259 & 0.2638 & 0.5381 & 0.5130 & 0.2833 & 0.4690 & 0.5570 \\
 & Ext. & GenBrain & 0.5218 & 0.2512 & 0.5165 & 0.4985 & 0.2594 & 0.4661 & 0.5309 \\
 & Ext. & MedicalNet & 0.5221 & 0.2263 & 0.5109 & 0.5134 & 0.2934 & 0.5175 & 0.5093 \\
 & Ext. & MRI-Core & 0.6292 & 0.3539 & \textbf{0.6338} & 0.5850 & 0.3482 & 0.5009 & \textbf{0.6692} \\
 & Ext. & RadFM & 0.5491 & 0.2770 & 0.5407 & 0.5194 & 0.2860 & 0.4830 & 0.5557 \\
\addlinespace[1pt]
4-shot & Ours & MAE P6 & \textbf{0.6916} & \textbf{0.4251} & \textbf{0.6492} & \textbf{0.6234} & \textbf{0.4043} & 0.5788 & \textbf{0.6681} \\
 & Ext. & BrainiAC & 0.5670 & 0.2773 & 0.5429 & 0.5440 & 0.3107 & 0.5458 & 0.5422 \\
 & Ext. & GenBrain & 0.5460 & 0.2712 & 0.5413 & 0.5281 & 0.2992 & 0.5054 & 0.5508 \\
 & Ext. & MedicalNet & 0.5390 & 0.2411 & 0.5231 & 0.5383 & 0.3151 & 0.5657 & 0.5108 \\
 & Ext. & MRI-Core & 0.6784 & 0.4052 & 0.6411 & 0.6214 & 0.4012 & \textbf{0.5855} & 0.6573 \\
 & Ext. & RadFM & 0.5859 & 0.3025 & 0.6030 & 0.5494 & 0.3089 & 0.4586 & 0.6402 \\
\addlinespace[1pt]
8-shot & Ours & JEPA P7 & \textbf{0.7386} & \textbf{0.4805} & \textbf{0.6788} & \textbf{0.6596} & \textbf{0.4419} & \textbf{0.6266} & \textbf{0.6926} \\
 & Ext. & BrainiAC & 0.5767 & 0.2920 & 0.5627 & 0.5550 & 0.3283 & 0.5416 & 0.5685 \\
 & Ext. & GenBrain & 0.5599 & 0.2761 & 0.5461 & 0.5373 & 0.3074 & 0.5206 & 0.5541 \\
 & Ext. & MedicalNet & 0.5789 & 0.2667 & 0.5299 & 0.5621 & 0.3512 & 0.6170 & 0.5072 \\
 & Ext. & MRI-Core & 0.7159 & 0.4518 & 0.6715 & 0.6430 & 0.4404 & 0.5946 & 0.6914 \\
 & Ext. & RadFM & 0.5863 & 0.3073 & 0.5704 & 0.5590 & 0.3385 & 0.5403 & 0.5778 \\
\addlinespace[1pt]
16-shot & Ours & JEPA P6 & \textbf{0.7867} & \textbf{0.5325} & \textbf{0.7057} & \textbf{0.7063} & \textbf{0.5031} & \textbf{0.7067} & 0.7059 \\
 & Ext. & BrainiAC & 0.6049 & 0.3154 & 0.6085 & 0.5665 & 0.3226 & 0.4942 & 0.6388 \\
 & Ext. & GenBrain & 0.6311 & 0.3411 & 0.5885 & 0.5852 & 0.3506 & 0.5793 & 0.5912 \\
 & Ext. & MedicalNet & 0.6640 & 0.3442 & 0.6317 & 0.6193 & 0.3995 & 0.5980 & 0.6406 \\
 & Ext. & MRI-Core & 0.7699 & 0.5158 & 0.7032 & 0.6883 & 0.4904 & 0.6624 & \textbf{0.7142} \\
 & Ext. & RadFM & 0.6212 & 0.3361 & 0.6006 & 0.5703 & 0.3374 & 0.5185 & 0.6220 \\
\addlinespace[1pt]
32-shot & Ours & JEPA P2 & \textbf{0.8198} & 0.5803 & 0.7444 & \textbf{0.7303} & \textbf{0.5371} & \textbf{0.7055} & 0.7550 \\
 & Ext. & BrainiAC & 0.6568 & 0.3749 & 0.6375 & 0.5988 & 0.3731 & 0.5329 & 0.6647 \\
 & Ext. & GenBrain & 0.6606 & 0.3752 & 0.6580 & 0.5989 & 0.3668 & 0.4978 & 0.6999 \\
 & Ext. & MedicalNet & 0.7261 & 0.4294 & 0.6928 & 0.6671 & 0.4575 & 0.6226 & 0.7116 \\
 & Ext. & MRI-Core & 0.8006 & \textbf{0.5865} & \textbf{0.7697} & 0.7024 & 0.5237 & 0.5877 & \textbf{0.8171} \\
 & Ext. & RadFM & 0.6678 & 0.3765 & 0.6386 & 0.6042 & 0.3697 & 0.5475 & 0.6609 \\
\addlinespace[1pt]
Scratch full FT & Scratch & ViT-Scratch P6 & 0.7939 & 0.5446 & 0.7126 & 0.7102 & 0.5267 & 0.7074 & 0.7129 \\
\bottomrule
\end{tabular}
\caption{IDH baseline details. Ours is the best MAE/JEPA checkpoint within each setting, selected by AUROC; external baselines are listed separately.}
\label{tab:app-idh-baselines}

\end{table}

\begin{table}[!htbp]
\centering
\scriptsize
\setlength{\tabcolsep}{2.2pt}
\renewcommand{\arraystretch}{0.95}
\begin{tabular}{@{}lllrrrrrrr@{}}
\toprule
Setting & Type & Model & AUROC & AUPRC & Acc. & BAcc. & F1 & Sens. & Spec. \\
\midrule
Linear & Ours & JEPA P7 & \textbf{0.7533} & \textbf{0.7037} & \textbf{0.6979} & \textbf{0.6813} & \textbf{0.6012} & 0.5700 & \textbf{0.7926} \\
 & Ext. & BrainiAC & 0.5289 & 0.4658 & 0.5234 & 0.4983 & 0.3711 & 0.3300 & 0.6667 \\
 & Ext. & GenBrain & 0.5819 & 0.4931 & 0.5787 & 0.5698 & 0.4580 & 0.5100 & 0.6296 \\
 & Ext. & MedicalNet & 0.6844 & 0.6648 & 0.6213 & 0.6159 & 0.5617 & 0.5800 & 0.6519 \\
 & Ext. & MRI-Core & 0.6911 & 0.6053 & 0.6255 & 0.6157 & 0.5273 & 0.5500 & 0.6815 \\
 & Ext. & RadFM & 0.6378 & 0.5845 & 0.5830 & 0.5878 & 0.5367 & \textbf{0.6200} & 0.5556 \\
\addlinespace[1pt]
kNN & Ours & MAE P7 & \textbf{0.7178} & \textbf{0.6376} & \textbf{0.6298} & \textbf{0.6194} & 0.5252 & 0.5500 & \textbf{0.6889} \\
 & Ext. & BrainiAC & 0.5965 & 0.5139 & 0.5489 & 0.5322 & 0.4001 & 0.4200 & 0.6444 \\
 & Ext. & GenBrain & 0.5072 & 0.4406 & 0.5149 & 0.5104 & 0.4360 & 0.4800 & 0.5407 \\
 & Ext. & MedicalNet & 0.5987 & 0.5371 & 0.5787 & 0.5815 & 0.5394 & 0.6000 & 0.5630 \\
 & Ext. & MRI-Core & 0.6493 & 0.5611 & 0.5872 & 0.5759 & 0.4654 & 0.5000 & 0.6519 \\
 & Ext. & RadFM & 0.5678 & 0.5142 & 0.5660 & 0.5781 & \textbf{0.5625} & \textbf{0.6600} & 0.4963 \\
\addlinespace[1pt]
1-shot & Ours & MAE P7 & \textbf{0.5217} & 0.4595 & \textbf{0.5183} & \textbf{0.5246} & 0.4506 & 0.5670 & 0.4822 \\
 & Ext. & BrainiAC & 0.4979 & 0.4454 & 0.5047 & 0.4990 & 0.3918 & 0.4610 & 0.5370 \\
 & Ext. & GenBrain & 0.4960 & 0.4424 & 0.4991 & 0.5037 & 0.4225 & 0.5340 & 0.4733 \\
 & Ext. & MedicalNet & 0.5098 & \textbf{0.4606} & 0.5136 & 0.5227 & \textbf{0.4684} & \textbf{0.5840} & 0.4615 \\
 & Ext. & MRI-Core & 0.4760 & 0.4304 & 0.4983 & 0.4936 & 0.3836 & 0.4620 & 0.5252 \\
 & Ext. & RadFM & 0.4900 & 0.4360 & 0.5055 & 0.4919 & 0.3393 & 0.4000 & \textbf{0.5837} \\
\addlinespace[1pt]
2-shot & Ours & MAE P4 & \textbf{0.6067} & \textbf{0.5542} & \textbf{0.5715} & \textbf{0.5771} & \textbf{0.5211} & \textbf{0.6150} & 0.5393 \\
 & Ext. & BrainiAC & 0.5549 & 0.5138 & 0.5570 & 0.5429 & 0.4206 & 0.4480 & \textbf{0.6378} \\
 & Ext. & GenBrain & 0.5041 & 0.4755 & 0.4974 & 0.4970 & 0.4171 & 0.4940 & 0.5000 \\
 & Ext. & MedicalNet & 0.5320 & 0.4872 & 0.5204 & 0.5265 & 0.4756 & 0.5670 & 0.4859 \\
 & Ext. & MRI-Core & 0.5336 & 0.4981 & 0.5319 & 0.5223 & 0.4092 & 0.4580 & 0.5867 \\
 & Ext. & RadFM & 0.5156 & 0.4703 & 0.5174 & 0.5176 & 0.4499 & 0.5190 & 0.5163 \\
\addlinespace[1pt]
4-shot & Ours & MAE P7 & \textbf{0.6502} & \textbf{0.5781} & \textbf{0.5898} & \textbf{0.5937} & \textbf{0.5459} & \textbf{0.6200} & \textbf{0.5674} \\
 & Ext. & BrainiAC & 0.5116 & 0.4755 & 0.5051 & 0.5012 & 0.4094 & 0.4750 & 0.5274 \\
 & Ext. & GenBrain & 0.5087 & 0.4791 & 0.5055 & 0.5104 & 0.4541 & 0.5430 & 0.4778 \\
 & Ext. & MedicalNet & 0.5686 & 0.5227 & 0.5383 & 0.5453 & 0.4981 & 0.5920 & 0.4985 \\
 & Ext. & MRI-Core & 0.5250 & 0.5000 & 0.5191 & 0.5120 & 0.4295 & 0.4640 & 0.5600 \\
 & Ext. & RadFM & 0.5084 & 0.4786 & 0.5009 & 0.4984 & 0.4123 & 0.4820 & 0.5148 \\
\addlinespace[1pt]
8-shot & Ours & MAE P7 & \textbf{0.6514} & \textbf{0.5837} & \textbf{0.6009} & \textbf{0.5972} & \textbf{0.5208} & 0.5730 & \textbf{0.6215} \\
 & Ext. & BrainiAC & 0.5676 & 0.5194 & 0.5477 & 0.5446 & 0.4703 & 0.5240 & 0.5652 \\
 & Ext. & GenBrain & 0.5022 & 0.4699 & 0.5094 & 0.5089 & 0.4390 & 0.5060 & 0.5119 \\
 & Ext. & MedicalNet & 0.6024 & 0.5603 & 0.5757 & 0.5758 & 0.5152 & \textbf{0.5760} & 0.5756 \\
 & Ext. & MRI-Core & 0.5644 & 0.5215 & 0.5251 & 0.5308 & 0.4758 & 0.5690 & 0.4926 \\
 & Ext. & RadFM & 0.5669 & 0.5222 & 0.5498 & 0.5489 & 0.4782 & 0.5430 & 0.5548 \\
\addlinespace[1pt]
16-shot & Ours & MAE P7 & \textbf{0.7032} & \textbf{0.6265} & \textbf{0.6357} & \textbf{0.6373} & \textbf{0.5770} & 0.6480 & 0.6267 \\
 & Ext. & BrainiAC & 0.5728 & 0.5213 & 0.5506 & 0.5360 & 0.4329 & 0.4380 & \textbf{0.6341} \\
 & Ext. & GenBrain & 0.5153 & 0.4842 & 0.5017 & 0.5034 & 0.4353 & 0.5150 & 0.4919 \\
 & Ext. & MedicalNet & 0.6264 & 0.5692 & 0.5749 & 0.5845 & 0.5488 & \textbf{0.6490} & 0.5200 \\
 & Ext. & MRI-Core & 0.5877 & 0.5361 & 0.5502 & 0.5555 & 0.5109 & 0.5910 & 0.5200 \\
 & Ext. & RadFM & 0.5613 & 0.5094 & 0.5294 & 0.5362 & 0.4923 & 0.5820 & 0.4904 \\
\addlinespace[1pt]
32-shot & Ours & MAE P7 & \textbf{0.7205} & \textbf{0.6452} & \textbf{0.6426} & \textbf{0.6478} & \textbf{0.6060} & \textbf{0.6830} & 0.6126 \\
 & Ext. & BrainiAC & 0.5710 & 0.5181 & 0.5545 & 0.5530 & 0.4909 & 0.5430 & 0.5630 \\
 & Ext. & GenBrain & 0.5044 & 0.4833 & 0.5128 & 0.5075 & 0.4283 & 0.4720 & 0.5430 \\
 & Ext. & MedicalNet & 0.6722 & 0.6380 & 0.6204 & 0.6126 & 0.5365 & 0.5600 & \textbf{0.6652} \\
 & Ext. & MRI-Core & 0.6294 & 0.5664 & 0.5753 & 0.5748 & 0.5018 & 0.5710 & 0.5785 \\
 & Ext. & RadFM & 0.5932 & 0.5285 & 0.5523 & 0.5588 & 0.5244 & 0.6020 & 0.5156 \\
\addlinespace[1pt]
Scratch full FT & Scratch & ViT-Scratch P2 & 0.6656 & 0.6271 & 0.6128 & 0.6098 & 0.5364 & 0.5900 & 0.6296 \\
\bottomrule
\end{tabular}
\caption{MCI baseline details. Ours is the best MAE/JEPA checkpoint within each setting, selected by AUROC; external baselines are listed separately.}
\label{tab:app-mci-baselines}

\end{table}

\begin{table}[!htbp]
\centering
\scriptsize
\setlength{\tabcolsep}{4pt}
\renewcommand{\arraystretch}{0.95}
\begin{tabular}{@{}lllrrr@{}}
\toprule
Setting & Type & Model & MAE & $R^2$ & Pearson $r$ \\
\midrule
Linear & Ours & JEPA P3 & \textbf{2.7596} & \textbf{0.8854} & \textbf{0.9413} \\
 & Ext. & BrainiAC & 4.0524 & 0.7582 & 0.8715 \\
 & Ext. & GenBrain & 3.7665 & 0.7897 & 0.8889 \\
 & Ext. & MedicalNet & 3.6527 & 0.7942 & 0.8922 \\
 & Ext. & MRI-Core & 2.7604 & 0.8841 & 0.9405 \\
 & Ext. & RadFM & 3.3667 & 0.8307 & 0.9115 \\
\addlinespace[1pt]
kNN & Ours & JEPA P5 & \textbf{2.4676} & \textbf{0.9003} & \textbf{0.9489} \\
 & Ext. & BrainiAC & 3.1393 & 0.8160 & 0.9038 \\
 & Ext. & GenBrain & 3.2131 & 0.8163 & 0.9037 \\
 & Ext. & MedicalNet & 3.0605 & 0.8326 & 0.9127 \\
 & Ext. & MRI-Core & 2.5146 & 0.8959 & 0.9466 \\
 & Ext. & RadFM & 2.8056 & 0.8651 & 0.9302 \\
\addlinespace[1pt]
8-shot & Ours & MAE P3 & \textbf{6.0776} & \textbf{0.3688} & \textbf{0.6744} \\
 & Ext. & BrainiAC & 7.5086 & 0.1492 & 0.5494 \\
 & Ext. & GenBrain & 7.7536 & 0.0539 & 0.4697 \\
 & Ext. & MedicalNet & 25.7669 & -105.2466 & 0.0612 \\
 & Ext. & MRI-Core & 6.7488 & 0.2060 & 0.6407 \\
 & Ext. & RadFM & 7.2373 & 0.2446 & 0.5970 \\
\addlinespace[1pt]
16-shot & Ours & MAE P2 & \textbf{4.9034} & \textbf{0.6247} & 0.8054 \\
 & Ext. & BrainiAC & 6.1341 & 0.4546 & 0.7054 \\
 & Ext. & GenBrain & 6.6741 & 0.2931 & 0.5969 \\
 & Ext. & MedicalNet & 17.1616 & -123.8946 & 0.1219 \\
 & Ext. & MRI-Core & 4.9280 & 0.6212 & \textbf{0.8111} \\
 & Ext. & RadFM & 5.7896 & 0.5175 & 0.7573 \\
\addlinespace[1pt]
32-shot & Ours & MAE P6 & 4.1423 & 0.7336 & 0.8597 \\
 & Ext. & BrainiAC & 5.5202 & 0.5595 & 0.7597 \\
 & Ext. & GenBrain & 5.5072 & 0.5192 & 0.7338 \\
 & Ext. & MedicalNet & 9.0784 & -19.2254 & 0.2270 \\
 & Ext. & MRI-Core & \textbf{3.9286} & \textbf{0.7625} & \textbf{0.8765} \\
 & Ext. & RadFM & 5.1700 & 0.6172 & 0.8027 \\
\addlinespace[1pt]
64-shot & Ours & MAE P6 & 3.6447 & 0.7988 & 0.8945 \\
 & Ext. & BrainiAC & 4.7537 & 0.6722 & 0.8234 \\
 & Ext. & GenBrain & 4.8462 & 0.6314 & 0.7983 \\
 & Ext. & MedicalNet & 5.3743 & 0.2525 & 0.6577 \\
 & Ext. & MRI-Core & \textbf{3.4067} & \textbf{0.8224} & \textbf{0.9084} \\
 & Ext. & RadFM & 4.6524 & 0.6887 & 0.8351 \\
\addlinespace[1pt]
128-shot & Ours & MAE P5 & 3.4162 & 0.8281 & 0.9113 \\
 & Ext. & BrainiAC & 4.4779 & 0.7119 & 0.8454 \\
 & Ext. & GenBrain & 4.3823 & 0.7028 & 0.8399 \\
 & Ext. & MedicalNet & 4.4783 & 0.5905 & 0.7841 \\
 & Ext. & MRI-Core & \textbf{3.1644} & \textbf{0.8500} & \textbf{0.9229} \\
 & Ext. & RadFM & 4.2875 & 0.7373 & 0.8607 \\
\addlinespace[1pt]
Scratch full FT & Scratch & ViT-Scratch P5 & 2.6815 & 0.8723 & 0.9357 \\
\bottomrule
\end{tabular}
\caption{AGE baseline details. Ours is the best MAE/JEPA checkpoint within each setting, selected by MAE; external baselines are listed separately.}
\label{tab:app-age-baselines}

\end{table}

\begin{table}[!htbp]
\centering
\footnotesize
\setlength{\tabcolsep}{4.5pt}
\begin{tabular}{@{}llrrrrrr@{}}
\toprule
Type & Model & \multicolumn{2}{c}{GLI} & \multicolumn{2}{c}{PED} & \multicolumn{2}{c}{GLI/PED macro} \\
\cmidrule(lr){3-4}\cmidrule(lr){5-6}\cmidrule(l){7-8}
 & & Dice & IoU & Dice & IoU & Dice & IoU \\
\midrule
Ours & JEPA P2 & 0.8590 & 0.7624 & \textbf{0.7476} & \textbf{0.6252} & \textbf{0.8033} & 0.6938 \\
Ext. & BrainiAC & 0.7231 & 0.5870 & 0.5773 & 0.4368 & 0.6502 & 0.5119 \\
Ext. & GenBrain & 0.5405 & 0.4032 & 0.1889 & 0.1260 & 0.3647 & 0.2646 \\
Ext. & MedicalNet & 0.7379 & 0.6112 & 0.6097 & 0.4783 & 0.6738 & 0.5447 \\
Ext. & MRI-Core & \textbf{0.8657} & \textbf{0.7738} & 0.7211 & 0.6161 & 0.7934 & \textbf{0.6949} \\
Ext. & RadFM & 0.7004 & 0.5612 & 0.5299 & 0.3923 & 0.6151 & 0.4768 \\
Scratch & ViT-Scratch & 0.7979 & 0.6780 & 0.6844 & 0.5524 & 0.7411 & 0.6152 \\
\bottomrule
\end{tabular}
\caption{BraTS fixed-input transfer on GLI and PED. Dice and IoU are reported per task and as an unweighted GLI/PED macro average. Inputs were already skull-stripped and registered, so this table is not a raw-data preprocessing ablation.}
\label{tab:app-brats-gli-ped-baselines}

\end{table}

\FloatBarrier

\subsection{P2-vs-Best Uncertainty Details}

Table~\ref{tab:p2-vs-best-uncertainty-full} reports all exploratory paired uncertainty comparisons between P2 and the empirically best P-level in each primary setting. Confidence intervals use paired bootstrap estimates, p-values use paired sign-flip tests, and BH q-values correct across the 24 comparisons. Unit denotes the paired unit used for the comparison: Fold pairs the five outer-fold scores; Sample pairs held-out AGE predictions; Case pairs held-out BraTS segmentation cases; Task-stratified case pairs BraTS cases within GLI/PED strata before macro aggregation; Same P-level indicates that P2 is already the empirical best level, so the paired difference is identically zero.

\begin{table}[!htbp]
\centering
\scriptsize
\setlength{\tabcolsep}{2pt}
\begin{tabular}{@{}L{0.18\textwidth}L{0.09\textwidth}lllL{0.18\textwidth}L{0.06\textwidth}L{0.13\textwidth}@{}}
\toprule
Setting & Comp. & P2 & Best & Improvement 95\% CI & p & q & Unit \\
\midrule
IDH LINEAR MAE & P2 vs P6 & 0.7966 & \textbf{0.8331} & +0.0366 [-0.0002, +0.0733] & 0.1875 & 0.5000 & Fold \\
IDH LINEAR JEPA & P2 vs P6 & 0.8652 & \textbf{0.8767} & +0.0115 [-0.0238, +0.0496] & 0.6875 & 0.9706 & Fold \\
IDH KNN MAE & P2 vs P5 & 0.7213 & \textbf{0.8058} & +0.0845 [+0.0668, +0.1050] & 0.0625 & 0.3000 & Fold \\
IDH KNN JEPA & P2 vs P6 & 0.8207 & \textbf{0.8511} & +0.0304 [-0.0101, +0.0645] & 0.1875 & 0.5000 & Fold \\
IDH FEWSHOT32 MAE & P2 vs P6 & 0.7127 & \textbf{0.7960} & +0.0833 [+0.0602, +0.1073] & 0.0625 & 0.3000 & Fold \\
IDH FEWSHOT32 JEPA & P2 vs P2 & \textbf{0.8198} & \textbf{0.8198} & +0.0000 [+0.0000, +0.0000] & 1.0000 & 1.0000 & Same P-level \\
MCI LINEAR MAE & P2 vs P7 & 0.6789 & \textbf{0.7456} & +0.0667 [+0.0178, +0.1267] & 0.1250 & 0.4297 & Fold \\
MCI LINEAR JEPA & P2 vs P7 & 0.6756 & \textbf{0.7533} & +0.0778 [-0.0370, +0.2093] & 0.3750 & 0.6923 & Fold \\
MCI KNN MAE & P2 vs P7 & 0.6374 & \textbf{0.7178} & +0.0804 [+0.0483, +0.1165] & 0.0625 & 0.3000 & Fold \\
MCI KNN JEPA & P2 vs P7 & 0.6950 & \textbf{0.7141} & +0.0191 [-0.0148, +0.0454] & 0.3750 & 0.6923 & Fold \\
MCI FEWSHOT32 MAE & P2 vs P7 & 0.6361 & \textbf{0.7205} & +0.0843 [+0.0459, +0.1388] & 0.0625 & 0.3000 & Fold \\
MCI FEWSHOT32 JEPA & P2 vs P7 & 0.6591 & \textbf{0.6897} & +0.0305 [-0.0208, +0.0853] & 0.4375 & 0.7500 & Fold \\
AGE LINEAR MAE & P2 vs P2 & \textbf{2.8836} & \textbf{2.8836} & +0.0000 [+0.0000, +0.0000] & 1.0000 & 1.0000 & Same P-level \\
AGE LINEAR JEPA & P2 vs P3 & 2.9800 & \textbf{2.7596} & +0.2173 [+0.1402, +0.2928] & $<0.0001$ & 0.0012 & Sample \\
AGE KNN MAE & P2 vs P2 & \textbf{2.4897} & \textbf{2.4897} & +0.0000 [+0.0000, +0.0000] & 1.0000 & 1.0000 & Same P-level \\
AGE KNN JEPA & P2 vs P5 & 2.4691 & \textbf{2.4676} & +0.0018 [-0.0379, +0.0422] & 0.9295 & 1.0000 & Sample \\
AGE FEWSHOT128 MAE & P2 vs P5 & 3.4795 & \textbf{3.4162} & +0.0645 [+0.0079, +0.1211] & 0.2500 & 0.6000 & Fold \\
AGE FEWSHOT128 JEPA & P2 vs P4 & 3.6463 & \textbf{3.6067} & +0.0409 [-0.0611, +0.1476] & 0.6875 & 0.9706 & Fold \\
BRATS-GLI MAE & P2 vs P4 & 0.8409 & \textbf{0.8439} & +0.0030 [-0.0006, +0.0070] & 0.1253 & 0.4297 & Case \\
BRATS-GLI JEPA & P2 vs P2 & \textbf{0.8590} & \textbf{0.8590} & +0.0000 [+0.0000, +0.0000] & 1.0000 & 1.0000 & Same P-level \\
BRATS-PED MAE & P2 vs P5 & 0.7103 & \textbf{0.7163} & +0.0060 [-0.0048, +0.0183] & 0.3476 & 0.6923 & Case \\
BRATS-PED JEPA & P2 vs P2 & \textbf{0.7476} & \textbf{0.7476} & +0.0000 [+0.0000, +0.0000] & 1.0000 & 1.0000 & Same P-level \\
BRATS GLI+PED macro MAE & P2 vs P5 & 0.7756 & \textbf{0.7776} & +0.0020 [-0.0036, +0.0083] & 0.5442 & 0.8708 & Task-stratified case \\
BRATS GLI+PED macro JEPA & P2 vs P2 & \textbf{0.8033} & \textbf{0.8033} & +0.0000 [+0.0000, +0.0000] & 1.0000 & 1.0000 & Same P-level \\
\bottomrule
\end{tabular}
\caption{Full P2-vs-best paired uncertainty results reconstructed from raw downstream outputs. Positive improvement means that the empirically best P-level improves over P2; for AGE MAE, positive values indicate error reduction. Unit specifies the pairing granularity: Fold uses outer-fold scores, Sample uses held-out AGE samples, Case uses held-out BraTS cases, Task-stratified case pairs BraTS cases within GLI/PED strata before macro aggregation, and Same P-level means P2 is already the empirical best level.}
\label{tab:p2-vs-best-uncertainty-full}

\end{table}

\FloatBarrier

\subsection{From-Scratch P-Level Control}

To test whether the observed P-level effects are merely artifacts of MAE or JEPA, we also trained randomly initialized ViT encoders end-to-end on labeled downstream data for the three case-level tasks. This control uses the same downstream splits and feasible P-levels as the pretrained comparisons, but removes self-supervised pretraining from the pipeline.

\begin{figure}[!htbp]
\centering
\includegraphics[width=0.68\textwidth]{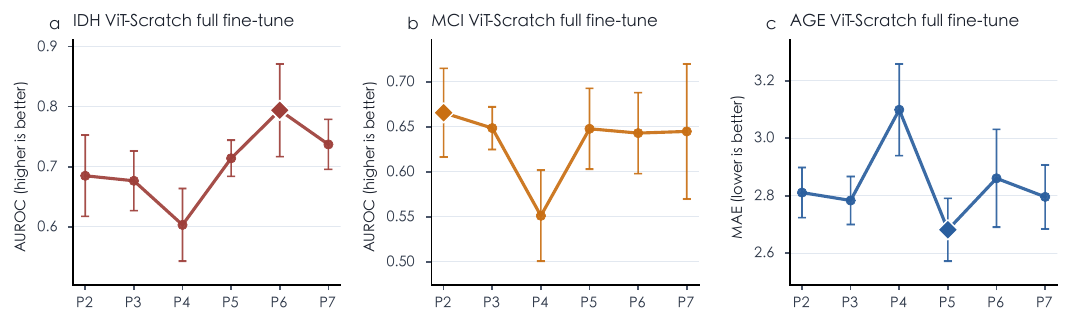}
\caption{From-scratch full fine-tuning control across feasible P-levels for the three case-level downstream tasks. Markers show five-fold means, error bars show fold-level standard deviation, and diamonds mark the best P-level within each task.}
\label{fig:from-scratch-control}
\end{figure}

Figure~\ref{fig:from-scratch-control} shows that preprocessing level still changes downstream behavior without self-supervised pretraining. In IDH, scratch full fine-tuning improves from 0.6849 AUROC at P2 to 0.7939 at P6, indicating that stronger spatial standardization can help even when the encoder is learned only from downstream labels. AGE shows a different optimum: the lowest MAE is obtained at P5 (2.6815), whereas P4 is substantially worse (3.0991). MCI does not follow the same pattern as the pretrained MCI results. Its best scratch AUROC is obtained at P2 (0.6656), and the heavier P-levels fluctuate rather than producing a monotonic gain.

These controls support a narrower but important conclusion. P-level is not only a technical detail of self-supervised pretraining; it is an input-distribution variable that can also affect fully supervised downstream training. At the same time, the scratch optima are task-dependent and do not uniformly favor heavier preprocessing. The from-scratch experiment should therefore be read as supporting evidence for the broader cost--utility framing, while the main evidence about foundation-model preprocessing remains the controlled MAE/JEPA sweep.

\FloatBarrier

\subsection{MCI Cross-Level Transfer Details}

Table~\ref{tab:app-mci-cross-level} provides the additional metrics for the MCI cross-level transfer experiment using the same organization as the main table. P2/P7 downstream data are shown as columns, and the off-diagonal delta is computed against the same-checkpoint matched-reference cell: for a P2 checkpoint, the delta compares P7 downstream data against P2 downstream data; for a P7 checkpoint, the delta compares P2 downstream data against P7 downstream data. Table~\ref{tab:app-mci-cross-level-mci-removed} reports the corresponding AUROC analysis after removing MCI-overlapping samples from the pretraining corpus.

\begin{table}[!htbp]
\centering
\footnotesize
\setlength{\tabcolsep}{5pt}
\begin{tabular}{@{}lllcccc@{}}
\toprule
Objective & Metric & Checkpoint & P2 data & P7 data & Off-diag. $\Delta$ & Gap \\
\midrule
JEPA & AUROC & P2 ckpt & \textbf{0.6756} & 0.7289 & +0.0533 & 68.6\% \\
JEPA & AUROC & P7 ckpt & 0.6378 & \textbf{0.7533} & -0.1156 & -- \\
JEPA & AUPRC & P2 ckpt & \textbf{0.6316} & 0.6226 & -0.0090 & -- \\
JEPA & AUPRC & P7 ckpt & 0.6019 & \textbf{0.7037} & -0.1019 & -- \\
JEPA & Acc. & P2 ckpt & 0.5915 & 0.6809 & +0.0894 & -- \\
JEPA & Acc. & P7 ckpt & \textbf{0.6255} & \textbf{0.6979} & -0.0723 & -- \\
JEPA & BAcc. & P2 ckpt & 0.5887 & 0.6807 & +0.0920 & -- \\
JEPA & BAcc. & P7 ckpt & \textbf{0.6144} & \textbf{0.6813} & -0.0669 & -- \\
JEPA & F1 & P2 ckpt & \textbf{0.5382} & \textbf{0.6260} & +0.0878 & -- \\
JEPA & F1 & P7 ckpt & 0.5306 & 0.6012 & -0.0706 & -- \\
\addlinespace[1pt]
MAE & AUROC & P2 ckpt & \textbf{0.6789} & 0.7330 & +0.0541 & 81.1\% \\
MAE & AUROC & P7 ckpt & 0.6574 & \textbf{0.7456} & -0.0881 & -- \\
MAE & AUPRC & P2 ckpt & \textbf{0.5979} & \textbf{0.6541} & +0.0562 & -- \\
MAE & AUPRC & P7 ckpt & 0.5948 & 0.6507 & -0.0560 & -- \\
MAE & Acc. & P2 ckpt & \textbf{0.6553} & \textbf{0.6894} & +0.0340 & -- \\
MAE & Acc. & P7 ckpt & 0.6000 & 0.6851 & -0.0851 & -- \\
MAE & BAcc. & P2 ckpt & \textbf{0.6611} & 0.6830 & +0.0219 & -- \\
MAE & BAcc. & P7 ckpt & 0.5948 & \textbf{0.6831} & -0.0883 & -- \\
MAE & F1 & P2 ckpt & \textbf{0.6389} & 0.6334 & -0.0055 & -- \\
MAE & F1 & P7 ckpt & 0.5156 & \textbf{0.6362} & -0.1206 & -- \\
\bottomrule
\end{tabular}
\caption{Additional MCI cross-level transfer metrics under linear probing. Entries follow the same organization as the main table: P2/P7 data are columns, and the off-diagonal $\Delta$ is computed relative to the same-checkpoint matched-reference cell.}
\label{tab:app-mci-cross-level}

\end{table}

\begin{table}[!htbp]
\centering
\footnotesize
\setlength{\tabcolsep}{5pt}
\begin{tabular}{@{}llcccc@{}}
\toprule
Objective & Checkpoint & P2 data & P7 data & Off-diag. $\Delta$ & Gap closure \\
\midrule
JEPA & P2 ckpt & 0.6638 & 0.7210 & +0.0572 & 173.9\% \\
JEPA & P7 ckpt & 0.6050 & 0.6967 & -0.0918 & -- \\
\addlinespace[1pt]
MAE & P2 ckpt & 0.6262 & 0.7607 & +0.1345 & 106.4\% \\
MAE & P7 ckpt & 0.6593 & 0.7526 & -0.0933 & -- \\
\bottomrule
\end{tabular}
\caption{MCI cross-level transfer after removing MCI-overlapping samples from the self-supervised pretraining corpus. Entries are linear-probe AUROC. Off-diagonal $\Delta$ is computed relative to the same-checkpoint matched-reference cell. Gap closure uses the matched P7-checkpoint/P7-data cell as the reference, following the same convention as the main-paper cross-level table.}
\label{tab:app-mci-cross-level-mci-removed}

\end{table}

\end{document}